%% file: pais.tex
\documentclass{ecai} 

\usepackage{latexsym}
\usepackage{amssymb}
\usepackage{amsmath}
\usepackage{amsthm}
\usepackage{booktabs}
\usepackage{enumitem}
\usepackage{graphicx}
\usepackage{color}
\usepackage{appendix}

\newcommand{\BibTeX}{B\kern-.05em{\sc i\kern-.025em b}\kern-.08em\TeX}

\begin{document}


\begin{frontmatter}


\paperid{66} 


\title{Automated Conversion of Static to Dynamic Scheduler via Natural Language}


\author[A]{\fnms{Paul Mingzheng}~\snm{TANG}}
\author[A]{\fnms{Kenji Kah Hoe}~\snm{LEONG}}
\author[A]{\fnms{Nowshad}~\snm{SHAIK}}
\author[A]{\fnms{Hoong Chuin}~\snm{LAU}\thanks{Corresponding Author}}

\address[A]{School of Computing and Information Systems, Singapore Management University}


\begin{abstract}
In this paper, we explore the potential application of Large Language Models (LLMs) that will automatically model constraints and generate code for dynamic scheduling problems given an existing static model. Static scheduling problems are modelled and coded by optimization experts. These models may be easily obsoleted as the underlying constraints may need to be fine-tuned in order to reflect changes in the scheduling rules. Furthermore, it may be necessary to turn a static model into a dynamic one in order to cope with disturbances in the environment.  In this paper, we propose a Retrieval-Augmented Generation (RAG) based LLM model to automate the process of implementing constraints for Dynamic Scheduling (\textsc{RAGDyS}), without seeking help from an optimization modeling expert. Our framework aims to minimize technical complexities related to mathematical modelling and computational workload for end-users, thereby allowing end-users to quickly obtain a new schedule close to the original schedule with changes reflected by natural language constraint descriptions.
\end{abstract}

\end{frontmatter}


\input{1_intro}


\input{2_related_work}


\input{3_proposed_method}


\input{4_evaluation}


\input{5_conclusion_future_work}


\input{6_acknowledgements}



\bibliography{mybibfile}

\clearpage
\appendix
\appendixpage
\input{appendix_gsp_nsp}

\input{appendix_planning_agent}
\input{appendix_coding_agent}

\end{document}

%% file: 1_intro.tex
\section{Introduction}

Companies today are facing increasing pressure to apply optimization models to solve complex resource planning and scheduling problems. Traditionally, these models are developed by human expert modelers and once they have been handed over and deployed in a production system, the models will remain unchanged until the next release of the system. 

In an increasing dynamic and uncertain world, the models developed may need to constantly evolve in order to remain relevant. Maintaining a team of human experts to keep the underlying optimization models updated is usually costly; many end-users organizations lament that the models expire quickly, and often they lack the expertise to keep the models updated in a production setting. 

Our research is motivated by the recent advances in large language models (LLM). Our goal is to develop a Generative AI methodology that enables the deployed resource optimization models to be updated (finetuned) by end-users themselves to cope with dynamics and uncertainty. This is a significant departure from the traditional modus operandi of hefty investments in either maintaining optimization modeling experts in-house or outsourcing the software maintenance to a third-party provider. In a wider context, this work is by the ideal to have human-AI collaboration via Generative AI in the context of resource planning, scheduling and optimization. While LLMs may not be directly relied upon for generating complex planning and scheduling models from ground zero \cite{LLMsstillcantplan}, we investigate in this paper the idea of prompt engineering for revising an existing planning model, which we believe to be a less daunting task. 

In this paper, we propose a LLM-based methodology that enables an existing static optimization model to be revised automatically in order to cope with dynamics and uncertainty. More precisely, we consider a well-known manpower scheduling/rostering problem, and explore how an LLM can be exploited to provide the necessary changes to the underlying constraints to cope with a dynamic variant of the underlying static scheduling problem. We assume that a static model exists and solutions for solving the model can be obtained via an efficient mathematical programming solver (such as Gurobi, Google OR-Tools etc). The interesting question we like to address is, without relying on a human modeling expert, to what extent can a planner interact with a pre-trained LLM to effectively generate the revised optimization model based on the static model? 


\subsection{Motivating Example}
To illustrate how a static nurse scheduling problem is extended into a dynamic scheduling problem, we present the following problem setting as a motivating example.

The static problem is to generate a schedule for nurses over a planning horizon over set of shift patterns. The following constraints for the schedule are:
 \begin{itemize}
     \item Every day, each shift is assigned to a single nurse.
     \item No nurse works more than one shift each day.
     \item Each nurse is assigned to at least M number of shifts overall.
 \end{itemize}
 
Each nurse has specified which shifts per day are preferable. The objective is to maximize the total number of preferences fulfilled.

We define the parameters below
\begin{itemize}
    \item $N$: Number of nurses
    \item $D$: Number of days
    \item $S$: Number of shifts
    \item $M$: Minimum number of shifts per nurse
    \item $P_{n,d,s}$: A boolean matrix over ($N,D,S$), indicating if nurse $n$ prefers to be assigned day $d$ and shift $s$.
\end{itemize}

We define the decision variables as:

$X_{n,d,s}$: An array of binary decision variables over ($N,D,S$), indicating if nurse $n$ is assigned to day $d$ and shift $s$.

We formulate the constraints below:
\begin{itemize}
    \item $\sum_{n}^{N} X_{n,d,s} = 1, \forall (d,s) \in D,S$
    \item $\sum_{s}^{S} X_{n,d,s} \leq 1, \forall (n,d) \in N,D$
    \item $\sum_{d}^{D} \sum_{s}^{S} X_{n,d,s} \geq M, \forall n \in N$
\end{itemize}

The objective is to maximise:
\begin{equation}
    \sum_{n}^{N} \sum_{d}^{D} \sum_{s}^{S} X_{n,d,s} \times P_{n,d,s}
\end{equation}

Now suppose that the model has been formulated and implemented in code written by an optimization expert. The end-user planner runs the code and obtains an initial schedule to be used, $X$. Now suppose on the day before the schedule was to be used, a nurse $A$ has to take a medical leave from day $D_1$ to $D_2$. We term this as a $disturbance$. The planner wants to ensure that the repaired schedule is within $T$ number of changes to $X$. We term $T$ as the perturbation threshold. In order for the planner to repair the schedule due to this and other disturbances, ideally the model need to be adjusted to include the following constraints:

\begin{equation}
    X_{A,d,s} = 0, \forall s \in S, \forall d \in [D_1, D_2]
\end{equation}
\begin{equation}
    \sum_{n}^{N} \sum_{d}^{D} \sum_{s}^{S} | X_{n,d,s} - X'_{n,d,s} | \leq T 
\end{equation}

Our proposed method will allow the planner to generate a revised model and the corresponding code to reflect these constraints, via a natural language description of the disturbance and perturbation. An example prompt chain for the planner could be: "Add a constraint such that nurse A is not available from day D1 to D2.". After generating the new code, the additional prompt is "Add a constraint such that the schedule generated does not change too much from the original schedule. The number of changes to the schedule should not exceed T."

\subsection{Contributions}
Our contributions are as follows:
\begin{itemize}
    \item We introduce the problem arising from developing such an automated tool for commercial manpower scheduling application in service organizations (such as restaurant chains and hospitals);
    \item  We propose a framework that enables the underlying constraint  model and the resulting code to be revised over time. This involves designing a dataset comprising natural language descriptions, mathematical formulations, and code samples both before and after incorporating dynamic constraints. 
    \item We utilize Retrieval-Augmented Generation (RAG) to generate the revised constraint formulation to solve the Minimum Perturbation Problem, whose aim is make use of an existing schedule obtained previously to generate a revised schedule automatically, such that the perturbation from the existing schedule is no more than the perturbation threshold $T$.
\end{itemize}

%% file: 2_related_work.tex
\section{Background and Related Work}
Before going through the details of our proposed method, we would like to introduce some main concepts and related work to contextualize our work: Constraint Programming and Optimization, Minimum Perturbation in Dynamic Scheduling, LLMs for Constraint Generation, and Retrieval-Augmented Generation.

\subsection{Constraint Programming and Optimization}
Constraint programming is now a relatively mature technology for solving a wide range of difficult combinatorial search problems \cite{ConstraintProgramming}, and in our context used to formulate our static and dynamic scheduling problems. Unlike Mixed-Integer Programming and Linear Programming models, Constraint programming models include functions to model more complex relations between variables, such as the \verb|AllDifferent| function. Due to addition of these functions, which are closer to natural language as opposed to linear constraints, we opted the use constraint programming formulations and code for our LLM planning and coding agents to learn from.

\subsection{Minimum Perturbation in Dynamic Scheduling}
The idea of modifying an existing scheduling problem (formulated as a Constraint Satisfaction Problem (CSP)) and the initial schedule to reflect new constraints introduced by disturbances, and maintaining a "minimally changed" new schedule, is termed a \textit{Minimum Perturbation Problem} (MPP)\cite{MPP1998}. The work formally defines a minimum perturbation problem as a 5-tuple ($\Theta_{i}$, $\alpha_{i}$, $C_{del}$, $C_{add}$, $\delta$).

\begin{itemize}
    \item $\Theta_{i}$ is a CSP
    \item $\alpha_{i}$ is a solution to $\Theta_{i}$
    \item $C_{del}$, $C_{add}$ are constraint removal and addition sets
    \item $\delta$ is a distance function between 2 complete solutions.
\end{itemize}

The original MPP defines a CSP, rather than a constraint optimization problem with an objective function. The objective function for the MPP is then defined as a minimization of the distance function $\delta$, which is described as as $d_x = |x - c|$ where $x$ is the new schedule variables and $c$ is the initial schedule assignments. For our problem, we formulate the MPP constraint as a hard constraint, where $d_x \leq T$ and $T$ is a threshold value for the maximum Hamming distance between the new and initial schedules. 
This minimum perturbation constraint serves 2 purposes: 1) It reduces the amount of unnecessary changes to the current schedule, which end-users are likely to want to be as stable as possible. 2) It reduces the search space considerably as we set a threshold for the maximum amount of change from the pre-computed schedule to the new schedule.

\subsection{LLMs for Constraint Generation}
There has been recent progress on using LLMs for combinatorial optimization, specifically on how to generate constraints for these problems based on natural language. For our work, we applied some elements from the OptiMUS framework presented in \cite{OptiMUS} such as the planning and coding agents, which breaks down the challenging problem of formulating constraints from natural language into smaller stages. Our work, however, differs in 1) the problem representation and 2) the context-specific constraint generation produced by the coding agent. To cater the the non-technical end-users of our framework, we do not expect them to formulate their problems in a structured form, and instead simply expects simple natural language to describe the new constraint to be added. To provide better context for the planning and coding agents, our scenario of dynamic scheduling already assumes an initial problem description, formulation, and code. By using these components, in addition to following including RAG examples, the planning agent is able to make use of existing parameters, variables, and constraints to generate a plan for new parameters, variables and constraints. The coding agent then makes use of the original code and the planning agent output to produce a modified code which implements the new constraint.

\subsection{Retrieval-Augmented Generation}
With minimizing technical expertise and computational resource requirements for end-users as our goal, Retrieval-Augmented Generation (RAG) techniques can be applied. RAG is commonly used to introduce context-specific knowledge which the pre-trained model may not have been trained on, as shown in \cite{RAG2020}, without the need for further fine-tuning. In the case of dynamic scheduling, worker scheduling problems can vary significantly and have context specific constraints which pre-trained LLMs are unlikely to be trained on. While there are various RAG techniques, as surveyed in \cite{RAGsurvey}, we opted to use Naive RAG for our method which has minimal fine-tuning steps.

\subsection{Challenges in Dynamic Constraint Generation}
For our problem of dynamic constraint generation, it would be preferable to have a set of benchmark worker scheduling problems to use, accompanied with natural language descriptions for each problem. There are a few challenges.

1) There is a lack of worker scheduling benchmark datasets available publicly. Most papers define the constraints specific to their problem and leave out the values of parameters required to construct a full problem instance to run.

2) There is a lack of natural language descriptions for worker scheduling problem instances. When scraping natural language description from optimization papers describing a worker scheduling problem, most papers use technical terms or mathematical formulation to express what they mean, which would differ significantly for an end-user's layperson description of a dynamic constraint to be added.

%% file: 3_proposed_method.tex
\section{Proposed Method}

\begin{figure}[ht]
\centering
\includegraphics[width=0.5\textwidth]{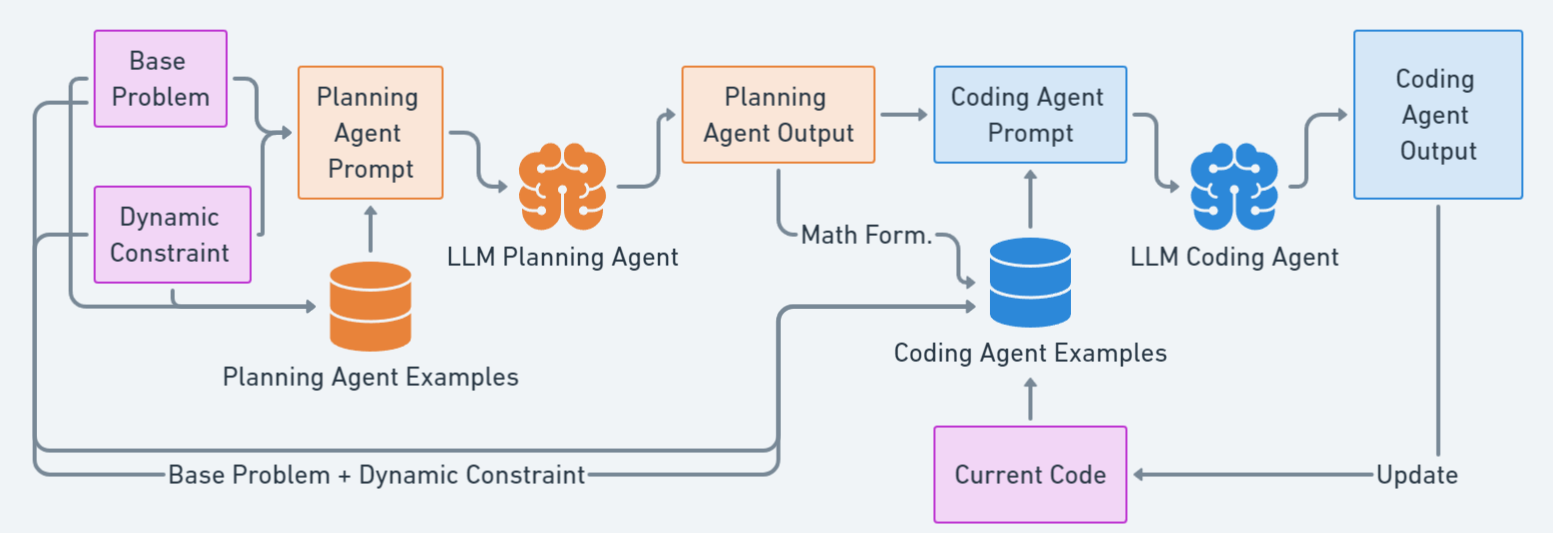}
\caption{Overview of our approach.}
\label{fig:overview}
\end{figure}

We introduce our proposed framework, \textsc{RAGDyS}, which takes as input (see pink components in Figure \ref{fig:overview}) a base problem description, a dynamic constraint description, and the current code for the base problem. In our context, the initial base problem description and current code are provided by an optimization expert, while the dynamic constraint description is provided by a non-technical end-user. The final output to be used would be the coding agent output, which is a modified version of the current code that includes code for new parameters, variables and constraints to reflect the dynamic constraint inserted. This new code can be executed to obtain a new schedule for the end-user. Further details of our approach and implementation are presented in the following section \ref{approach_details}.

\subsection{Architecture and Approach Details} \label{approach_details}
\textsc{RAGDyS} contains 3 stages. In the planning and coding stages, we use RAG methods to inform the LLM agents with the most similar example input output pair. The vector database used for our experiments was implemented using ChromaDB, which by default creates document embedding vectors using all-MiniLM-L6-v2 \cite{minilm}, and performs cosine similarity search to find the most similar example prompts from ChromaDB database.

\subsubsection{Step 1: Planning Stage}

\begin{figure}[ht]
\centering
\includegraphics[width=0.5\textwidth]{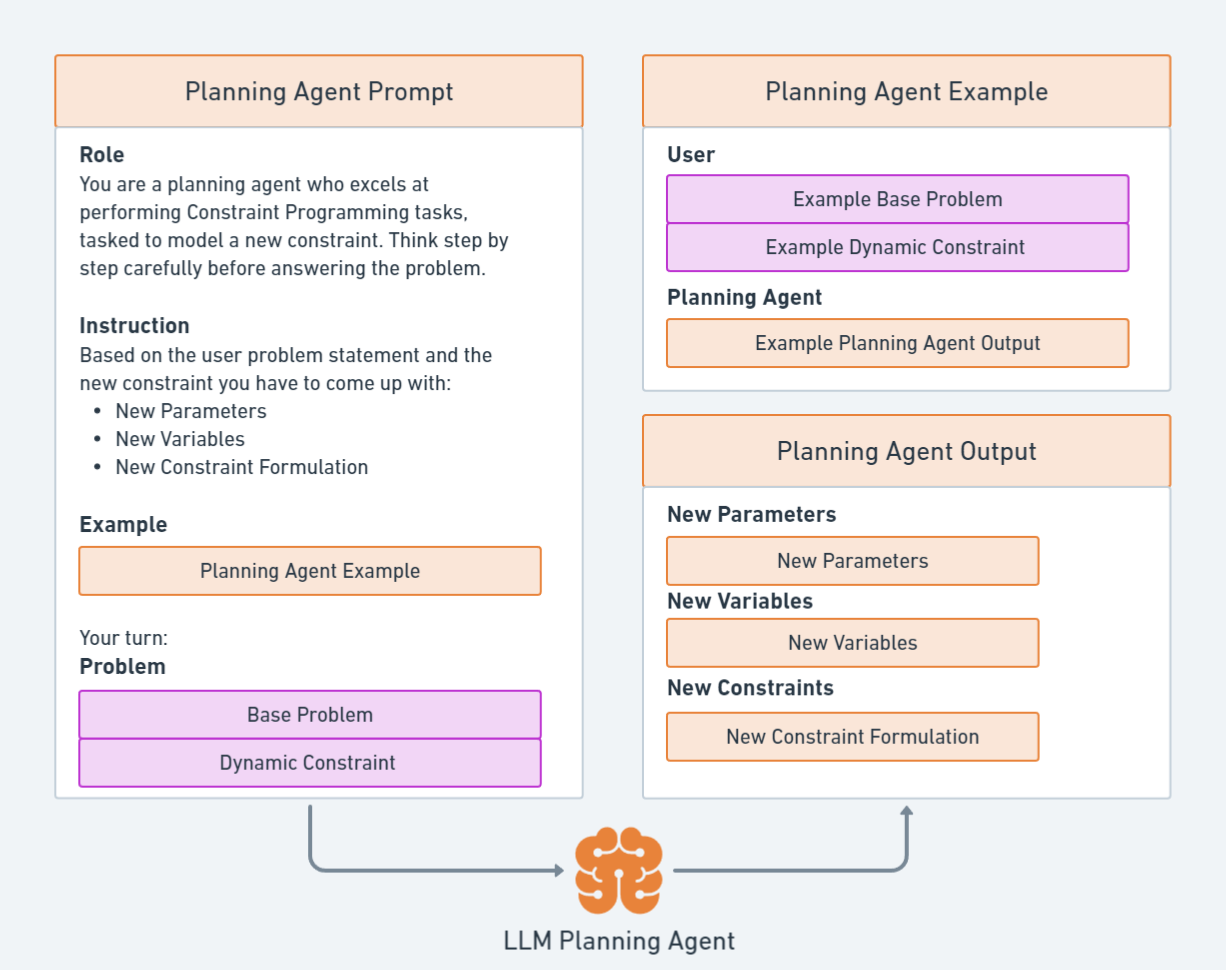}
\caption{Planning Stage Details.}
\label{fig:planning_agent}
\end{figure}

The planning stage takes as input the base problem description and dynamic constraint description. Following RAG methods, the most similar example is retrieved from a vector database of planning agent examples, which contain example user input and planning agent output pairs. These examples, along with the base problem and dynamic constraint descriptions, are inserted into the planning agent prompt template (See Appendix \ref{appendix_plan_example} for example planning agent prompt). The planning agent prompt is sent to an LLM agent (our implementation uses Gemini \cite{Gemini}), and the output of the LLM agent should contain 3 sections: New parameters, New variables and New constraints. See Figure \ref{fig:planning_agent} for more details. This output is then used in the coding stage.

\subsubsection{Step 2: Coding Stage}

\begin{figure}[ht]
\centering
\includegraphics[width=\columnwidth]{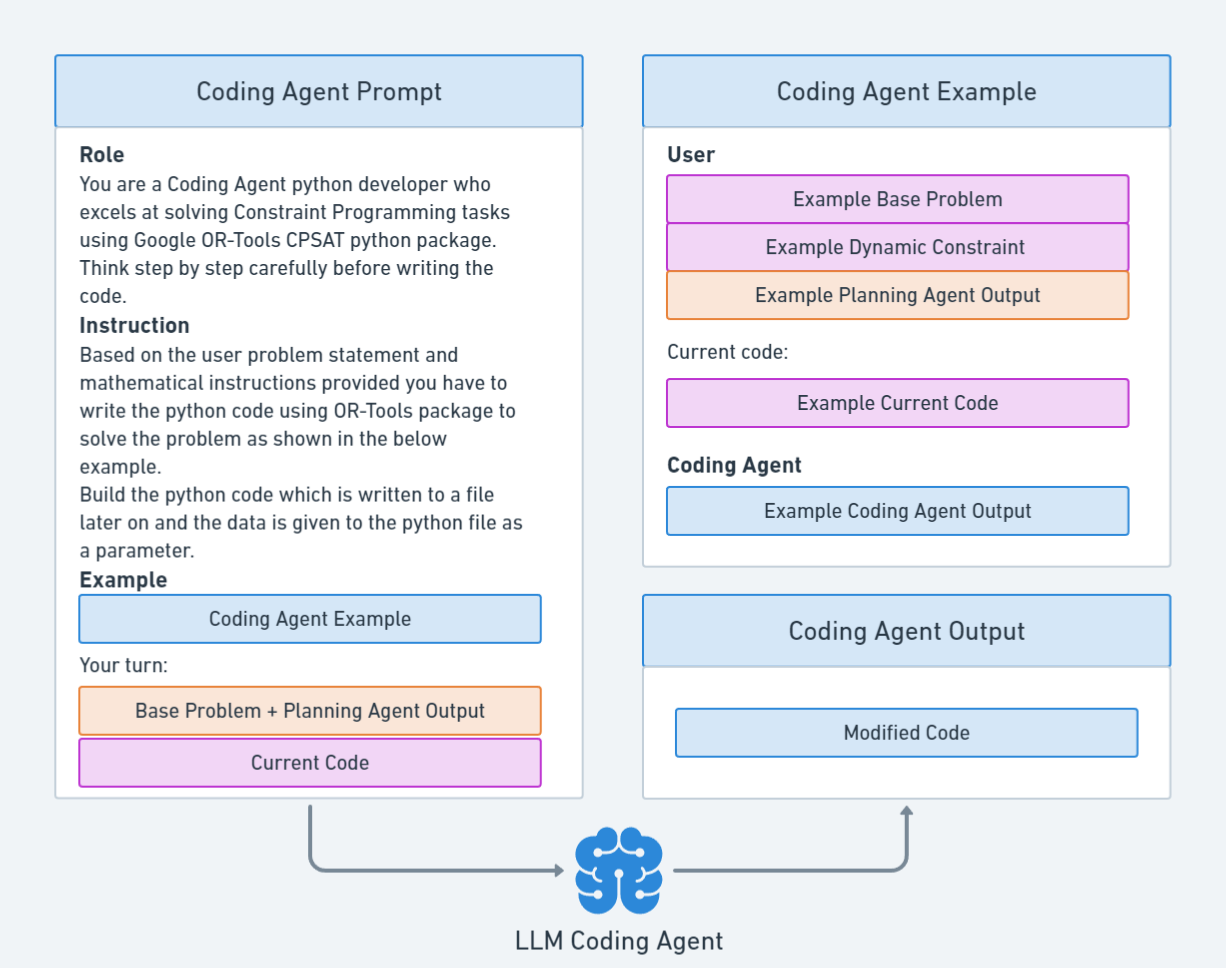}
\caption{Coding Stage Details.}
\label{fig:coding_agent}
\end{figure}

The coding stage takes as input the base problem and dynamic constraint descriptions, the planning agent output, and the current code to be modified. Similar to the planning stage, we obtain most similar example user input and coding agent output pairs. All these components are inserted into the coding agent prompt template (See Appendix \ref{appendix_code_example} for example coding agent prompt) to produce the coding agent prompt. The LLM coding agent (implemented with Claude Haiku \cite{Claude} and OpenAI GPT-4 \cite{GPT}), will take this prompt as input and produce the modified version of the code as output. See Figure \ref{fig:coding_agent} for more details. 

\subsubsection{Step 3: Execution Stage}

\begin{figure}[ht]
\centering
\includegraphics[width=\columnwidth]{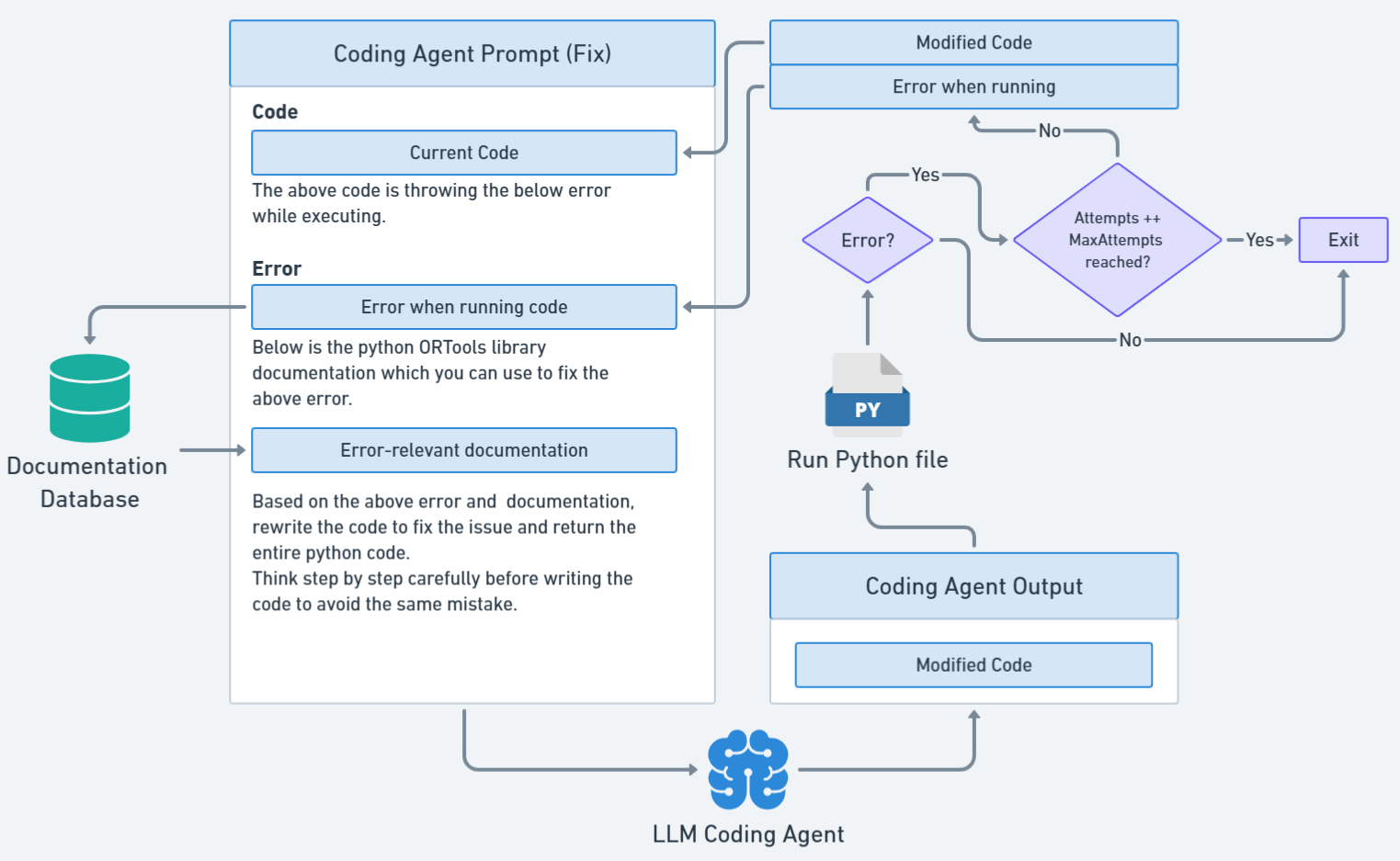}
\caption{Code-fixing stage Details.}
\label{fig:code-fixing}
\end{figure}

The execution stage is where the user is able to decide how to progress from the generated code output from the coding stage. The user may choose to either execute the modified code to produce a new schedule, or repeat the process to add another dynamic constraint. Our framework proposes to not only add the dynamic constraint associated with a disturbance, but also leveraging the existing schedule (if it exists) and minimal adjustments to obtain a new schedule efficiently through a repeated process of adding the minimum perturbation constraint using natural language.

When executing the new code, the code may produce errors and our framework will enters the code-fixing stage. In the code-fixing stage, the coding engine is called repeatedly to fix the code. The number of attempts cannot exceed the \verb|maxAttempts| threshold. Using RAG, we retrieve the most relevant document stored in a database containing documentation on Google OR-Tools 
CP-SAT objects and methods. Combining the error message, relevant documentation, and current code, we prompt the coding engine to amend the current code to fix the error. See Figure \ref{fig:code-fixing} for overall pipeline on the code-fixing stage.

\subsection{Dataset Generation} \label{dataset_gen}

\begin{figure}[ht]
\centering
\includegraphics[width=\columnwidth]{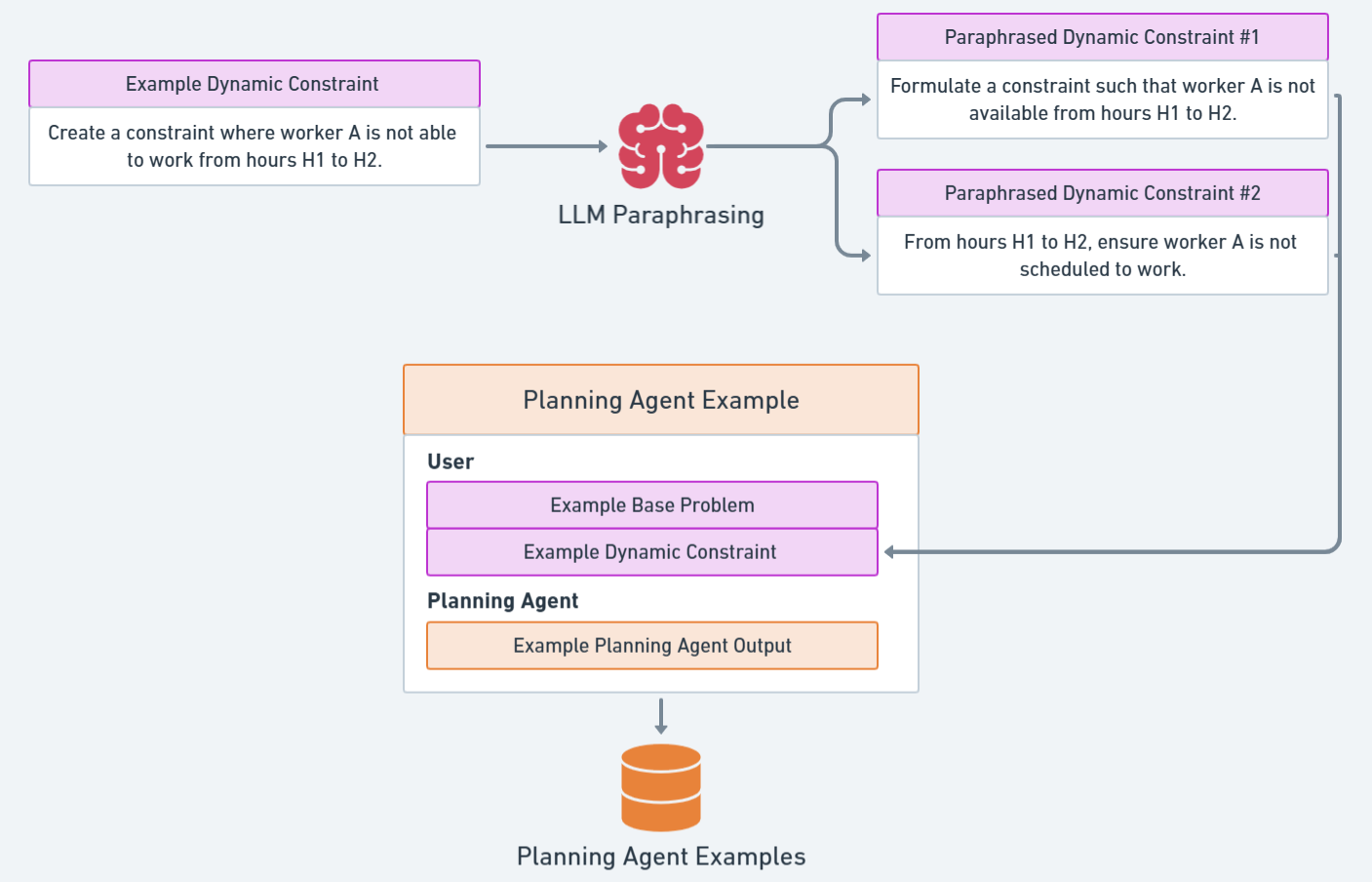}
\caption{LLM paraphrasing for dataset generation}
\label{fig:llm-paraphrase}
\end{figure}

We manually prepare a data set to be inserted into the vector database and used as references for RAG when solving unseen problems. We also manually write a test set for the unseen problem instances. We multiply the total number of instances used in our database and testing set by prompting a LLM to generate paraphrased versions of the natural language descriptions, as shown in Figure \ref{fig:llm-paraphrase}. Due to possible mistakes generated when the LLM paraphrases the description (removes or changes the name of parameters to be used in the constraint), we opt to manually fix these mistakes.

\subsubsection{Database Set}
We manually write natural language prompts with a base worker scheduling problem (In our case, we only use the Gig-Scheduling Problem (GSP)), planning agent outputs, and before and after constraint insertion codes written in Python using the Google OR-Tools CP-SAT solver library.

GSP is a variant of the worker scheduling problem, where workers are not assigned a shift out of a fixed set of shifts per day, but are assigned directly to hours subject to constraints(such as specifying the minimum and maximum length of consecutive hours assigned). These worker-hour assignments must then match, as much as possible, the task demands. Tasks are defined as a job to be assigned to a worker for a single hour, and is constrained by skill requirements (only a subset of workers with the qualified skills can be assigned to the task), and no multi-tasking (a worker can only be assigned at most 1 task per hour). See Appendix \ref{appendix_gsp} for the mathematical programming formulations for the base GSP.

\subsubsection{Test Set}
Similar to the Database set, we manually write natural language prompts, planning agent outputs, and code for the GSP. In addition, to test if our proposed method is able to generalize to different base problems, we also manually construct data for the standard Nurse Scheduling Problem (NSP). See Appendix \ref{appendix_nsp} for the mathematical programming formulations for the base NSP. We use the same LLM paraphrasing method to increase the initial testing set of size 14 to size 70 (35 GSP instances, 35 NSP instances).

%% file: 4_evaluation.tex
\section{Evaluation and Results}

In the following, we report experimental results. Besides performing automatic evaluation, we also conduct manual inspection to evaluate the correctness of the constraints and code. As described in \ref{dataset_gen}, we populated the dataset using GSP problem instances, and test the performance of the planning and coding agents using 2 sets: 1) Unseen dynamic constraints for the GSP, to verify if the LLMs are able to generalise dynamic constraints with similar base problem); and 2) Unseen dynamic constraints for the NSP, to verify if the LLMs are able to generalise to different base worker scheduling problem and apply dynamic constraints to the new context.

\subsection{Automatic Evaluation}
Similar to OptiMUS, we can evaluate the performance of our model by executing the generated code from the coding agent and comparing the results with the expected results. There are a few possible outcomes when running the code:
\begin{itemize}
    \item KeyError for data
    \item Syntax error in expressing constraints
    \item Infeasible solution
    \item Feasible solution but objective value does not match
    \item Solution matches expectation
\end{itemize}

We classify the outcome as a solution matching the target when one of the following are met:
\begin{itemize}
    \item If both target and generated code reach an optimal solution, then the 2 solutions are exactly equal.
    \item If both target and generated code are at least feasible, then check if the computed upper and lower bounds are similar (measure the overlap between the 2 intervals). If the overlap covers half the total coverage, it is considered a match.
\end{itemize}

We should note that the results of the automatic evaluation method are not completely reliable and produce false positives (Detect a Match in expected solution, but the constraint formulated differs from what the prompt intends) and false negatives (Detect a mismatch in solution despite correct constraint formulation. This is due to the non-deterministic nature of running Google OR-Tools with multi-threading). Hence, we also conduct manual evaluation to investigate these possible outcomes, as well as look provide more in-depth explanations for the errors.

\subsection{Manual Evaluation}
During manual inspection, we delve deeper into the specifics behind the results displayed in the automatic evaluation. This includes identifying the reasons behind the outcomes and double-checking solutions for both false positives and false negatives.

We can identify the following:
\begin{itemize}
    \item False Positives
    \item False Negatives
    \item Hallucinating Parameters
    \item Incorrect Reasoning
\end{itemize}

In the following section, we detail our experimental results.

\subsection{Results and Discussion}

\begin{table}[ht]
\resizebox{\columnwidth}{!}{%
\begin{tabular}{lcccccc}
\hline
      & \multicolumn{5}{c}{Outcome}                                                                                                                                             & \multicolumn{1}{l}{}      \\ \hline
      & \multicolumn{1}{l}{KeyError} & \multicolumn{1}{l}{Syntax Error} & \multicolumn{1}{l}{Infeasible} & \multicolumn{1}{l}{Feasible (Not Match)} & \multicolumn{1}{l}{Match} & \multicolumn{1}{l}{Total} \\ \hline
GSP   & 0                            & 0                                & 0                              & 6                                        & 29                        & 35                        \\ \hline
NSP   & 0                            & 1                                & 0                              & 0                                        & 34                        & 35                        \\ \hline
Total & 0                            & 1                                & 0                              & 6                                        & 63                        & 70                        \\ \hline
\end{tabular}%
}
\caption{Evaluation results on test sets using OpenAI GPT4}
\label{overall_results_automatic_openai}
\end{table}

Table \ref{overall_results_automatic_openai} shows the overall results for automatic evaluation with OpenAI GPT-4 as the coding agent. In general, automatic evaluation has shown that for both seen and unseen constraints for the same and different base problems, the model attains a 90\% match rate. As mentioned in the earlier section, we would need to further inspect the output code for the cases of false positives and false negatives. In the next section, we measure the false positive and negatives, as well as analyze the code outputs to obtain explanations for the errors. 

\begin{table}[ht]
\resizebox{\columnwidth}{!}{%
\begin{tabular}{l|c|c|c|c|}
\cline{2-5}
                          & \multicolumn{1}{l|}{False Positives} & \multicolumn{1}{l|}{False Negative} & \multicolumn{1}{l|}{Parameter Hallucination} & \multicolumn{1}{l|}{Incorrect Reasoning} \\ \hline
\multicolumn{1}{|l|}{GSP} & 6                                    & 2                                   & 0                                            & 9                                        \\ \hline
\multicolumn{1}{|l|}{NSP} & 4                                    & 0                                   & 3                                            & 1                                        \\ \hline
\end{tabular}%
}
\caption{Manual Evaluation results on test sets OpenAI GPT4}
\label{manual_eval_nsp_table}
\end{table}

\subsubsection{Example False Positive}
\begin{figure}[ht]
\centering
\includegraphics[width=\columnwidth]{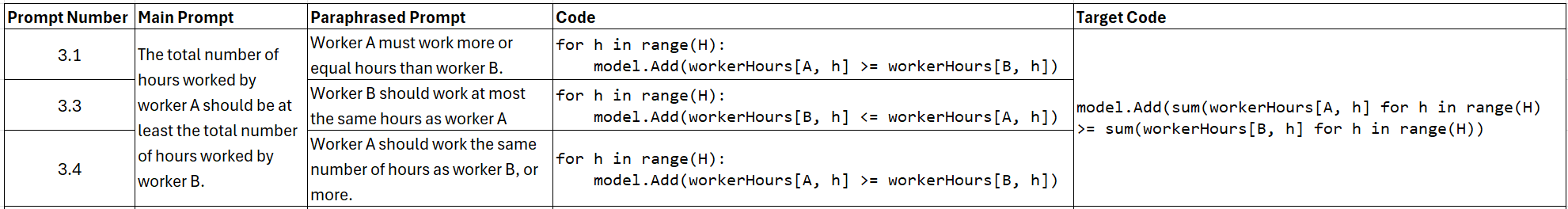}
\caption{Table of results for false positive cases for GSP prompt 3 and the paraphrased versions.}
\label{fig:false_pos_gsp}
\end{figure}

As shown in Figure \ref{fig:false_pos_gsp}, there are instances where the automatic evaluation flags the result as a match, but the code generated does not formulate the correct constraint. Looking at the code and target code, we can see that the coding agent has misinterpreted the paraphrased prompt as "Worker A must have at least the same schedule as Worker B, with additional hours added on" rather than "Total number of hours worked by worker A >= Total number of hours worked by worker B". It is possible that this error is caused by poor phrasing of the paraphrased prompt.

\subsubsection{Example False Negative}
\begin{figure}[ht]
\centering
\includegraphics[width=\columnwidth]{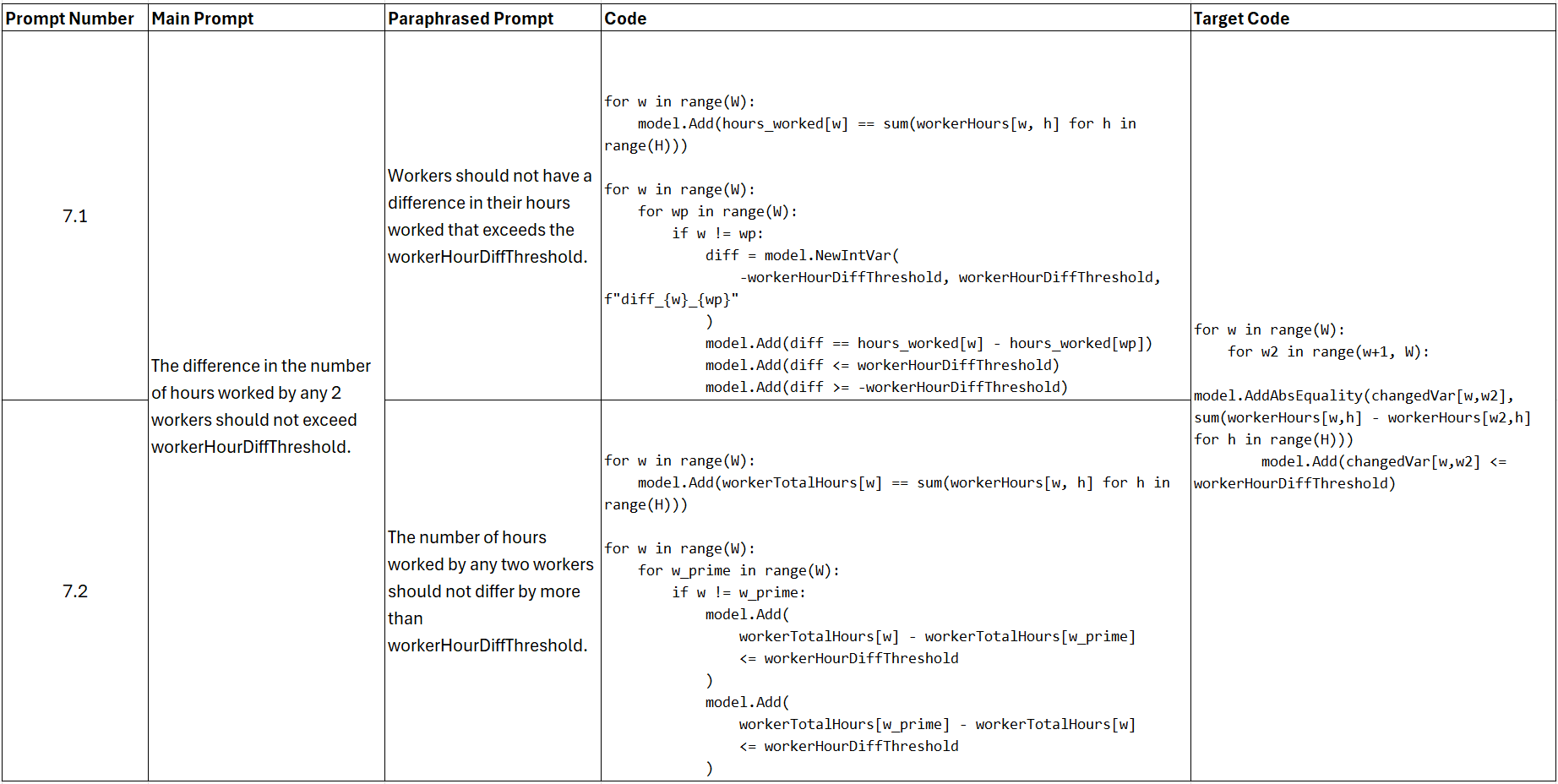}
\caption{Table of results for false negative cases for GSP prompt 7 and the paraphrased versions.}
\label{fig:false_neg_gsp}
\end{figure}

In Figure \ref{fig:false_neg_gsp}, we observe that the generated code is logical and is consistent with the target code, despite sometimes creating unnecessary variables or constraints. However, the automatic evaluation flagged the result as a mismatch, because the objective value obtained after running the code differs from the target code's objective value. This is a result of how ORTools handles different constraint formulations and how one formulation could be computationally more efficient, as well as the non-deterministic nature of ORTools when running on multiple threads, meaning even with the exact same code running there is a possibility that the output will differ if we do not allow the solver to run to optimality. Due to the nature of the GSP instance, we opted to run the code with multiple threads and a fixed time limit so as to obtain results in reasonable time.

\subsubsection{Example Parameter Hallucination}
\begin{figure}[ht]
\centering
\includegraphics[width=\columnwidth]{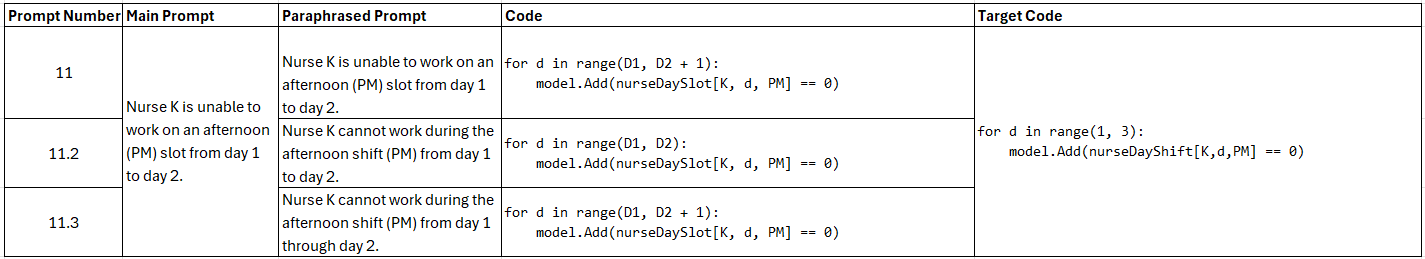}
\caption{Table of results for false positive cases for NSP prompt 11 and the paraphrased versions, which exemplify parameter hallucination}
\label{fig:halluc_param}
\end{figure}

As shown in Figure \ref{fig:halluc_param}, we can see an example of parameter hallucination when observing prompts like prompt 11. In this NSP prompt, we opted to use numbers directly instead of variable names for the days, such as D1 and D2. Because most code is written using the variable formulation, there is a possibility that the LLM coding agent hallucinates parameters for day 1 to day 2, replacing it with D1 and D2 as parameters expected to be provided by the user. 

\subsubsection{Example Incorrect Reasoning} 
\begin{figure}[ht]
\centering
\includegraphics[width=\columnwidth]{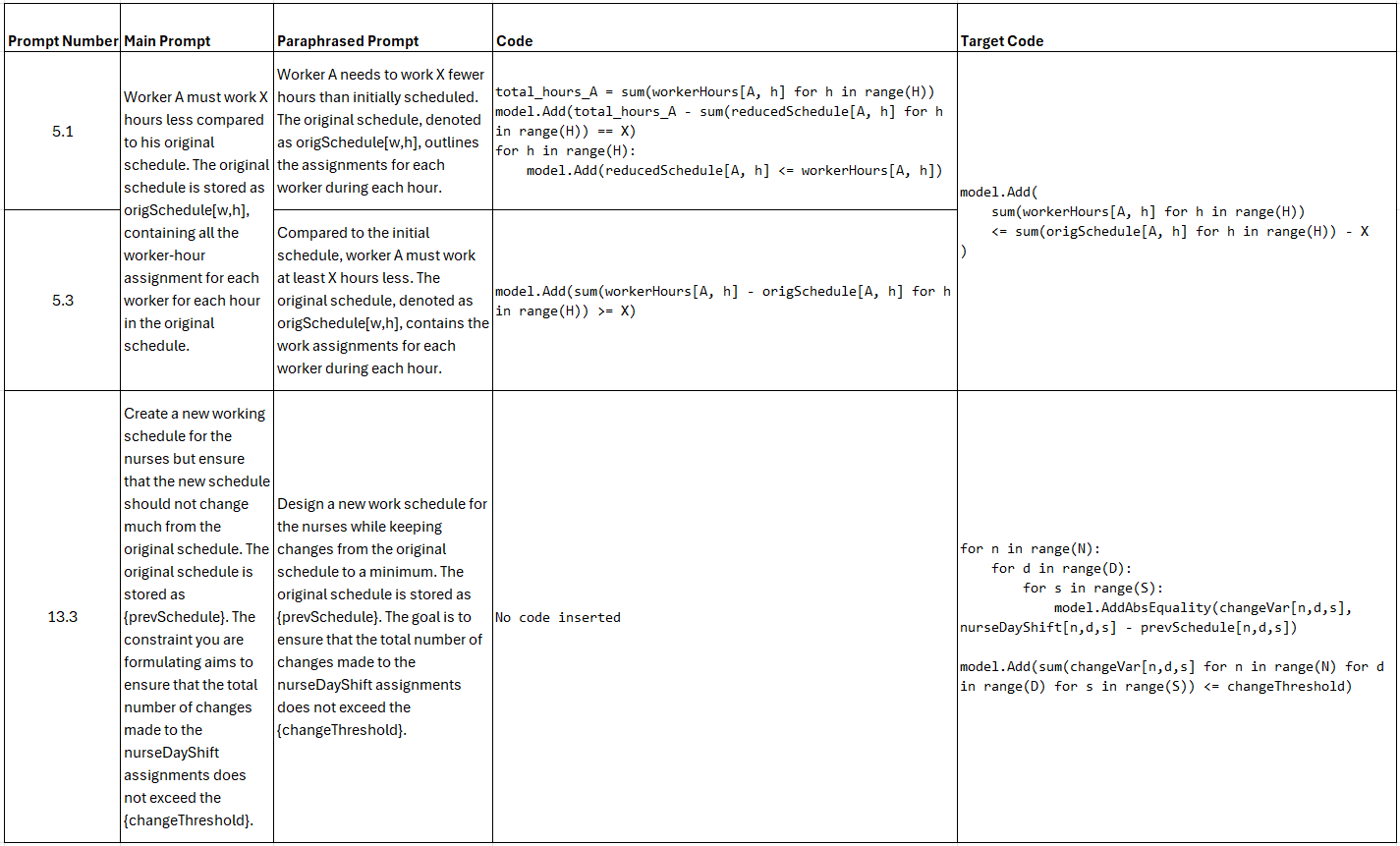}
\caption{Table of results for false positive cases where we identify incorrect reasoning causing incorrect / empty constraints being generated}
\label{fig:incor_reason}
\end{figure}

In Figure \ref{fig:incor_reason}, we can see a few examples of incorrect reasoning. For prompt 13.3, the LLM coding agent did not generate any additional code on top of the base code, implying that it was simply unable to reason what the prompt wanted (Minimum perturbation constraint). For prompts 5.1, we can see a combination of hallucinated parameters (reducedHours) and incorrect reasoning (not using origSchedule parameter at all, as well as adding illogical constraints). Lastly, in prompt 5.3 we can see an almost correct constraint, however the change in number of hours from origSchedule should be computed with the expression "origSchedule[A,h] - workerHours[A,h]" rather than "workerHours[A,h] - origSchedule[A,h]".

\subsection{Ablation Study}
We conduct an ablation study on how a different LLM (Claude Haiku) performs against OpenAI GPT-4 when being used as the coding agent. The results of the experiments are shown in Table \ref{overall_results_automatic_claude_haiku}.

\subsubsection{Ablation 1: Varying the LLM}
\begin{table}[ht]
\resizebox{\columnwidth}{!}{%
\begin{tabular}{lcccccc}
\hline
      & \multicolumn{5}{c}{Outcome}                                                                                                                                             & \multicolumn{1}{l}{}      \\ \hline
      & \multicolumn{1}{l}{KeyError} & \multicolumn{1}{l}{Syntax Error} & \multicolumn{1}{l}{Infeasible} & \multicolumn{1}{l}{Feasible (Not Match)} & \multicolumn{1}{l}{Match} & \multicolumn{1}{l}{Total} \\ \hline
GSP   & 5                            & 3                                & 1                              & 5                                        & 21                        & 35                        \\ \hline
NSP   & 7                            & 1                                & 5                              & 0                                        & 22                        & 35                        \\ \hline
Total & 12                            & 4                                & 6                              & 5                                        & 43                        & 70                        \\ \hline
\end{tabular}%
}
\caption{Evaluation results on test sets using Claude Haiku}
\label{overall_results_automatic_claude_haiku}
\end{table}

\begin{figure}[ht]
\centering
\includegraphics[width=\columnwidth]{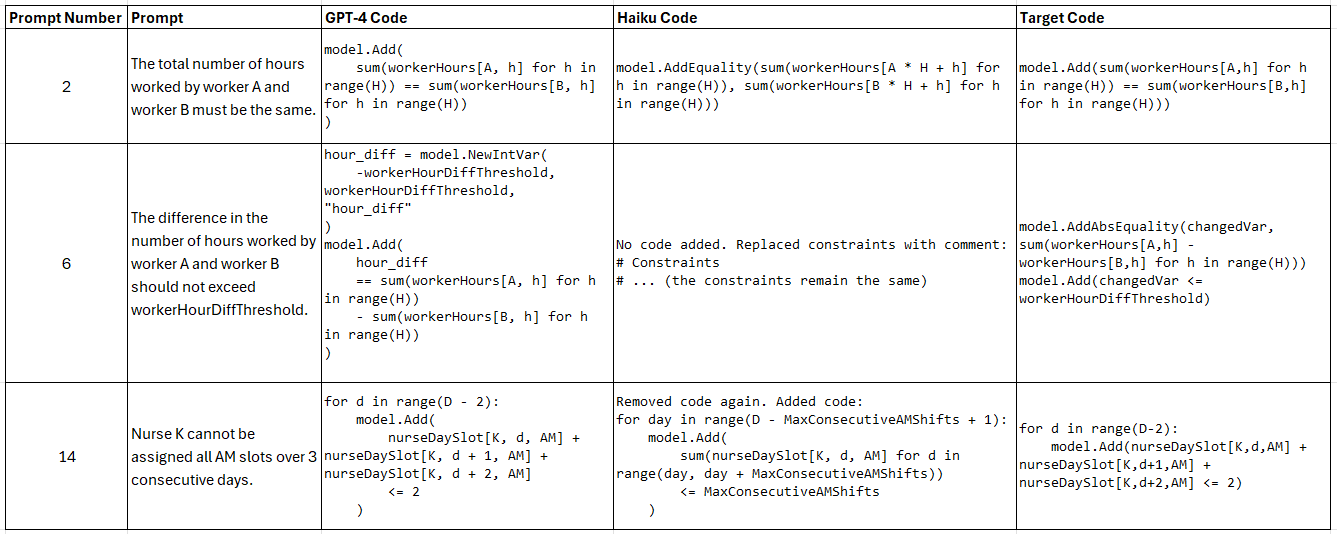}
\caption{Table of some prompts and generated code from OpenAI GPT-4 and Claude Haiku as the coding agent.}
\label{fig:haiku_compare}
\end{figure}

We can observe the difference in performance using GPT-4 and Claude Haiku, looking at Tables \ref{overall_results_automatic_openai} and \ref{overall_results_automatic_claude_haiku}. As expected of the more expensive model, OpenAI GPT-4 outperforms Claude Haiku when used in the coding agent, with an automatic evaluation match rate of 90\% while Claude Haiku has an automatic match rate of 60\%.

Looking at the outputs of Claude Haiku in Figure \ref{fig:haiku_compare}, we can see its limitations as opposed to GPT-4. Very often, Claude Haiku will remove the existing constraints despite instructions in the system prompt to output the full code. In addition, it frequently hallucinates parameters such as \verb|MaxConsecutiveAMShifts| and even library functions such as \verb|AddEquality|.

%% file: 5_conclusion_future_work.tex
\section{Conclusion and Future Work}
We propose a framework \textsc{RAGDyS} which automates the process of adding dynamic constraints to existing static scheduling problems, thereby minimizing the need for human optimization experts. 

For future work, we plan to encompass a wider range of worker scheduling problems and explore other combinatorial optimization problems that can be formulated with minimum perturbation constraints. One such example is dynamic pickup and delivery routing problems. Routing problems are well-studied and routing-specific methods (such as \verb|AddCircuit| in ORTools) in constraint programming libraries should be useful and intuitive for LLMs to make use of.
Additionally, further prompt optimization can be done, specifically on the intricacies and techniques specific to the solver library. For example, to further improve the performance of the coding agent, our coding agent prompt requires additional knowledge on how the Python absolute (abs) function cannot be used when defining variables and constraints in OR-Tools. We should explore strategies for prompt optimization and apply them to enhance the current framework.

%% file: 6_acknowledgements.tex
\begin{ack}
We would like to thank Nayan Rokhade, Derek Cho, and Ming Bing Ng for assistance during code implementation.
\end{ack}

%% file: appendix_gsp_nsp.tex
\section{Gig Scheduling Problem (GSP) Formulation}\label{appendix_gsp}

\subsection{Variables}
\begin{itemize}
    \item $W$, $H$, $T$ represents the number of workers, hours and tasks respectively. 
    \item $x_{w,h}$ and $y_{w,t}$ is the binary worker-hour assignment and the binary worker-task assignment.
    \item $A_{w,h}$ and $S_{w,t}$ is the worker-hour availability and worker-task skill compatibility parameters.
    \item $taskHour(t)$ returns the hour of task $t$, and $T_h$ is the set of tasks in hour $h$.
    \item $s_{w,h}$ and $e_{w,h}$ are the intermediate binary variables indicating the start and end of a block of consecutive working hours starting / ending at hour $h$ for worker $w$.
    \item $BMin$, $BMax$, and $RMin$ are the parameters for minimum block length, maximum block length, and minimum rest length.
    \item $u_t$ is the soft constraint variable indicating if task $t$ is unassigned.
\end{itemize}

\subsection{Constraint Description}
\begin{itemize}
    \item \textbf{Worker Hour Availability}: hard constraint ensuring that workers can only be assigned hours they are available in as specified in $A_{w,h}$.
    \item \textbf{Worker Task Skill Requirements}: hard constraint ensuring that workers can only be assigned tasks if they meet the skill requirements, specified in $workerTaskSkills[w,t]$. 
    \item \textbf{Worker Task Hour Precedence}: hard constraint ensuring that a worker must be assigned the task's hour before being assigned the task.
    \item \textbf{No Multi-task}: hard constraint ensuring that a worker can only work at most 1 task per hour. 
    \item \textbf{Minimum Block Length}: hard constraint ensuring an assigned block of hours for a worker is at least $BMin$ long.
    \item \textbf{Maximum Block Length}: hard constraint ensuring an assigned block of hours for a worker is at most $BMax$ long.
    \item \textbf{Minimum Rest}: hard constraint ensuring the length of rest time between consecutive blocks is at least $RMin$ long. 
    \item \textbf{Unassigned Tasks}: soft constraint which counts the number of tasks that are not assigned to any worker.
\end{itemize}

\subsection{Constraint Formulation}
\begin{equation}
    x_{w,h} \leq A_{w,h}, \forall w \in W, h \in H
\end{equation}
\begin{equation}
    y_{w,t} \leq S_{w,t}, \forall w \in W, t \in T
\end{equation}
\begin{equation}
    y_{w,t} \leq x_{w,taskHour(t)}, \forall w \in W, t \in T
\end{equation}
\begin{equation}
    \sum_{t}^{T_h} y_{w,t} \leq 1, \forall w \in W, \forall h \in H 
\end{equation}

Intermediate constraints formulating the start and end of blocks.
\begin{equation}
    s_{w,h} \geq x_{w,h} - x_{w,h-1}, \forall w \in W, \forall h \in H
\end{equation}
\begin{equation}
    s_{w,0} = x_{w,0}, \forall w \in W
\end{equation}
\begin{equation}
    e_{w,h} \geq x_{w,h} - x_{w,h+1}, \forall w \in W, \forall h \in H
\end{equation}
\begin{equation}
    e_{w,H-1} = x_{w,H-1}, \forall w \in W
\end{equation}

Minimum block length constraint
\begin{equation}
    \sum_{h' = h}^{h + BMin - 1} x_{w,h'} \geq s_{w,h} \times BMin
\end{equation}

Maximum block length constraint
\begin{equation}
    \sum_{h' = h}^{h + BMax} 1 - x_{w,h'} \geq s_{w,h}
\end{equation}

Minimum Rest constraint
\begin{equation}
    e_{w,h} \leq x_{w,h'}, \forall w \in W, \forall h \in H, \forall h' \in [h, h + RMin]
\end{equation}

Unassigned Task constraint
\begin{equation}
    \sum_{w}^{W} y_{w,t} \geq 1 - u_t, \forall t \in T
\end{equation}

\subsection{Objective}
Minimize the following:
\begin{equation}
      \sum_{w,h}^{W,H} x_{w,h} + \alpha \sum_{t}^{T} u_t
\end{equation}
$\alpha = 1000$

\section{Nurse Scheduling Problem (NSP) Formulation}\label{appendix_nsp}

\subsection{Variables}
\begin{itemize}
    \item $N$, $D$, $S$, $T$ represents the number of nurses, days and shifts and slots respectively.
    \item There are 3 slots, morning AM, afternoon PM, and night ND.
    \item $x_{n,d,s}$ and $y_{n,d,t}$ is the binary nurse-day-shift assignment and the binary nurse-day-slot assignment.
    \item $A_{n,d,s}$ is the nurse availability parameter for each day and shift.
    \item $shiftSlot(s)$ returns the slot number for shift $s$.
    \item $shiftHours(s)$ returns the number of hours for shift $s$.
    \item $Y_{d,t}$ is the demand for the nurses required on day $d$ slot $t$.
    \item $R_{n,d}$ is the specified rest days for each nurse for each day.
    \item $MinHours$, $MaxHours$, $MaxWorkDays$ represent the minimum hours worked per nurse, maximum hours worked per nurse, and the maximum days worked per nurse.
    \item $surplus$ and $shortfall$ are intermediate integer variables counting the over or under supply of nurses on day $d$ slot $t$.
\end{itemize}

\subsection{Constraint Description}
\begin{itemize}
    \item \textbf{Nurse Availability}: a nurse can only be assigned a day and shift if she is available.
    \item \textbf{At most 1 shift per day}: a nurse can at most be assigned 1 shift per day.
    \item \textbf{Nurse Shift Slot}: a nurse assigned to a shift indicates the nurse is assigned the shift's slot.
    \item \textbf{Demand constraint}: the number of nurses assigned to day $d$ and slot $t$ equals $Y_{d,t} + surplus_{d,t} - shortfall_{d,t}$
    \item \textbf{Minimum hours}: minimum hours assigned to nurse is $MinHours$.
    \item \textbf{Maximum hours}: maximum hours assigned to nurse is $MaxHours$.
    \item \textbf{Maximum days}: maximum days assigned to a nurse is $MaxDays$
    \item \textbf{No ND-AM pattern}: a nurse cannot work a night slot followed by a morning slot.
\end{itemize}

\subsection{Constraint Formulation}
\begin{equation}
    x_{n,d,s} \leq A_{n,d,s}, \forall n,d,s \in N,D,S
\end{equation}
\begin{equation}
    \sum_{s}^{S} x_{n,d,s} \leq 1, \forall n,d \in N,D
\end{equation}
\begin{equation}
    x_{n,d,s} \leq y_{n,d,shiftSlot(s)}, \forall n,d,s \in N,D,S
\end{equation}

\begin{equation}
    \sum_{d,s}^{D,S} x_{n,d,s} \times shiftHours(s) \geq MinHours, \forall n \in N
\end{equation}
\begin{equation}
    \sum_{d,s}^{D,S} x_{n,d,s} \times shiftHours(s) \leq MaxHours, \forall n \in N
\end{equation}
\begin{equation}
    \sum_{d,s}^{D,S} x_{n,d,s} \leq MaxDays, \forall n \in N
\end{equation}

Demand Constraint
\begin{equation}
    \sum_{n}^{N} y_{n,d,t} = Y_{d,t} + surplus_{d,t} - shortfall_{d,t}
\end{equation}

No ND-AM Pattern
\begin{equation}
    y_{n,d,ND} + y_{n,d+1,AM} \leq 1, \forall n,d \in N,D
\end{equation}

\subsection{Objective}
Minimize the following:
\begin{equation}
    \sum_{d,t}^{D,T} surplus_{d,t} + shortfall_{d,t}
\end{equation}

%% file: appendix_planning_agent.tex
\clearpage
\onecolumn
\section{Planning Agent Prompt Example}\label{appendix_plan_example}

\begin{verbatim}
### Role ###
You are a planning agent who excels at performing Constraint Programming tasks, tasked to model a new constraint.
Think step by step carefully before answering the problem.

### Instruction ###
Based on the user problem statement and the new constraint you have to come up with:
- New Parameters
- New Variables
- New Constraint Formulation

### Example ###
*User*
Problem:
You are tasked to solve a worker scheduling problem. 
In this problem, there are {W} workers, {H} hours, and {T} tasks.
Each worker can be assigned to each hour.
Each task has a fixed hour, and can be assigned to each worker. {taskHour[t]} specified which hour task t is assigned.
The main task is to assign workers to hours and tasks to workers, subject to various constraints.
The objective is to minimize the following:
total_hours_assigned + {UNASSIGNED_PENALTY} * total_number_of_unassigned_tasks

Parameters:
W: Number of workers
H: Number of hours
T: Number of tasks
BMin: Minimum block length
BMax: Maximum block length
RMin: Minimum rest length
availableHours[w,h]: Indicates if worker w is available to be assigned hour h (1 if available, otherwise 0)
workerTaskSkills[w,t]: Indicates if worker w is skilled enough to be assigned task t (1 if sufficiently skilled, otherwise 0)
taskHour[t]: Returns the hour of the task
UNASSIGNED_PENALTY: Penalty per unassigned task in the objective function

Decision Variables:
workerHours[w,h]: Indicates if worker w is assigned hour h
workerTasks[w,t]: Indicates if worker w is assigned task t
startBlock[w,h]: Indicates if worker w is assigned a consecutive block of hours starting at hour h
endBlock[w,h]: Indicates if worker w is assigned a consecutive block of hours ending at hour h
unassignedTasks[t]: Indicates if task t is not assigned all workers

Constraints:
1) Available hours constraint. If a worker is assigned an hour, he must be available for that hour.
workerHours[w,h] <= availableHours[w,h] for each w for each h
2) Worker task skill constraint. If a worker is assigned a task, he must be skilled enough for that task.
workerTask[w,t] <= workerTaskSkills[w,t] for each w for each t
3) Task hour assignment constraint. If a worker is assigned a task, he must be assigned the task's hour.
workerTask[w,t] <= workerHours[w,taskHour[t]] for each w for each t
4) At most 1 task assigned for each worker for each hour
sum(workerTask[w,t] for each t if taskHour[t]=h) <= 1 for each w for each h
5) Start block constraint. If a worker w starts a block of consecutive worker-hour assignments at hour h, startBlock[w,h] = 1. If workerHours[w,0], then startBlock[w,0]
workerHours[w,h] + (1-workerHours[w,h-1]) - 1 <= startBlock[w,h] for each w for each h
workerHours[w,0] <= startBlock[w,0] for each w
6) End block constraint. If a worker w ends a block of consecutive worker-hour assignments at hour h, endBlock[w,h] = 1. If workerHours[w,H-1], then endBlock[w,H-1]
workerHours[w,h] + (1-workerHours[w,h+1]) - 1 <= endBlock[w,h] for each w for each h
workerHours[w,H-1] <= endBlock[w,H-1] for each w
7) Minimum block length constraint. If a worker starts a block at hour h, the worker must be assigned the at least {BMin} hours.
startBlock[w,h] <= workerHours[w,x] for each w for each h for each x in range(h,h+BMin)
8) Maximum block length constraint. If a worker starts a block at hour h, the worker must end within {BMax} hours.
startBlock[w,h] <= sum(endBlock[w,x] for x in range(h,h+BMax)) for each w for each h
9) Minimum Rest constraint. The minimum length of rest (block of consecutive worker-hour = 0) between the end of a block and the start of a block is {RMin}.
endBlock[w,h] <= 1-workerHours[w,x] for each w for each h for each x in range(h+1,h+RMin+1)
10) Unassigned Task constraint. If a task is not assigned to any worker, unassignedTasks[t] = 1.

Given the above context, provide the mathematical formulation for the new constraint mentioned below.:
New Constraint:
11) Formulate a constraint such that worker A does not work from hours H1 to H2.

You are free to add parameters and decision variables.

New Parameters:
A: Worker affected from hours H1 to H2
H1: Start of hours where worker A does not work
H2: End of hours where worker A does not work

New Variables:
None

New Constraints:
workerHours[A,h] = 0 for each h in range(H1,H2)

Your turn
### Problem ###
You are tasked to solve a nurse rostering problem. 
In this problem, there are {N} nurses, {D} days, {S} shifts and {T} slots.
Each worker can be a shift for each day.
Each shift takes a certain number of hours, specified in {shiftHours}.
Each shift fills either a Morning(AM)(Slot 0), Afternoon(PM)(Slot 1), or Night(ND)(Slot 2) slot.
There is a demand requirement for each day for each slot, where we want to assign the nurses their shifts as close as possible to fulfil this demand.
The main task is to assign nurses a shift for each day (or no shift if the nurse is not working for that day), while satisfying some constraints and minimizing an objective.
The objective is to minimize the following:
surplus + shortfall
surplus is the total number of nurses assigned to a slot more than the specified demand for that day and slot.
shortfall is the total number of nurses assigned to a slot less than the specified demand for that day and slot.

Parameters:
N: Number of nurses
D: Number of days
S: Number of shifts
T: Number of slots
availableShifts[n,d,s]: Indicates if nurse n is available to be assigned day d and slot s (1 if available, otherwise 0)
shiftSlot[s] returns the slot number for shift s. (Eg. 7am to 11am is a Morning shift(AM), so it has a value of 0 (Slot 0))
shiftHours[s] = Integer value for the number of hours in shift s. (Eg. 7am to 11am shift has a value of 4)
demandSlot[d,t] = Integer value of the demand required specified per day d per slot t
restDays[n,d]: Indicates if nurse n is resting (Not available) on day d. 
MinHours = data["MinHours"]
MaxHours = data["MaxHours"]
MaxWorkingDays = data["MaxWorkingDays"]

Decision Variables:
nurseDayShift[n,d,s]: Indicates if nurse n is assigned day d and shift s
nurseDaySlot[n,d,t]: Indicates if nurse n is assigned a shift in day d slot t
surplus[d,t] = Integer value of the surplus nurses assigned on day d slot t. (If X more nurses are assigned to day d slot t than specified in demandSlot[d,t], return X, otherwise 0)
shortfall = Integer value of the shortfall nurses assigned on day d slot t. (If X less nurses are assigned to day d slot t than specified in demandSlot[d,t], return X, otherwise 0)

Constraints:
1) Available shift constraint. If nurse n is assigned day d and shift s, she must be available on day d shift s.
nurseDayShift[n,d,s] <= availableShifts[n,d,s] for each n for each d for each s
2) Max 1 shift per day constraint. At most 1 shift per day constraint
sum(nurseDayShift[n,d,s] for each s) <= 1, for each n for each s
3) Shift to slot assignment constriant. If a nurse n is assigned day d shift s, then nurseDaySlot[n,d,shiftSlot[s]] = 1
nurseDaySlot[n,d,t] <= sum(nurseDayShift[n,d,s] for each s if shiftSlot[s] == t)
4) Demand Requiremented Constraint. Number of nurses assigned for each day d for each slot t == demandSlot[d,t] + surplus[d,t] - shortfall[d,t]
sum(nurseDaySlot[n,d,t] for each n) == demandSlot[d,t] + surplus[d,t] - shortfall[d,t]
5) Minimum Hours Constraint. Each nurse must work at at least {MinHours}
sum(shiftHours[s] * nurseDayShift[n,d,s] for each d for each s) >= MinHours, for each n
6) Maximum Hours Constraint: Each nurse must work at most {MaxHours}
sum(shiftHours[s] * nurseDayShift[n,d,s] for each d for each s) <= MaxHours, for each n
7) Rest Days Constraint. If a nurse n is assigned any shift on day d, she must not be on a rest.
nurseDayShift[n,d,s] <= 1 - restDays[n,d] for each n for each d for each s
8) Maximum number of working days per week constraint. Each nurse must work at most (MaxWorkingDays) per week.
sum(nurseDayShift[n,w*7 + d,s] for d in range(7) for each s) <= MaxWorkingDays, for each n for each week w
9) No consecutive Night(ND) shift - Morning (AM) shift assignment for any nurse constraint.
nurseDaySlot[n,d,ND] + nurseDaySlot[n,d+1,AM] <= 1, for each n for each d

New Constraint
Nurse K cannot work on days D1.

\end{verbatim}

%% file: appendix_coding_agent.tex
\clearpage
\onecolumn
\begin{small}
\section{Coding Agent Prompt Example}\label{appendix_code_example}

\begin{verbatim}
### Role ###
You are a Coding Agent python developer who excels at solving Constraint Programming tasks using Google OR-Tools CPSAT python package.
Think step by step carefully before writing the code.

### Instruction ###
Based on the user problem statement and mathematical instructions provided you have to write the python code
using OR-Tools package to solve the problem as shown in the below example.
Build the python code which is written to a file later on and the data is given to the python file as a parameter.

### Note ###
Note that your response should only contain python code without any explanation or run command is needed.

### Example ###
*User*
Problem:
You are tasked to solve a worker scheduling problem. 
In this problem, there are {W} workers, {H} hours, and {T} tasks.
Each worker can be assigned to each hour.
Each task has a fixed hour, and can be assigned to each worker. {taskHour[t]} specified which hour task t is assigned.
The main task is to assign workers to hours and tasks to workers, subject to various constraints.
The objective is to minimize the following:
total_hours_assigned + {UNASSIGNED_PENALTY} * total_number_of_unassigned_tasks

Parameters:
W: Number of workers
H: Number of hours
T: Number of tasks
BMin: Minimum block length
BMax: Maximum block length
RMin: Minimum rest length
availableHours[w,h]: Indicates if worker w is available to be assigned hour h (1 if available, otherwise 0)
workerTaskSkills[w,t]: Indicates if worker w is skilled enough to be assigned task t (1 if sufficiently skilled, otherwise 0)
taskHour[t]: Returns the hour of the task
UNASSIGNED_PENALTY: Penalty per unassigned task in the objective function

Decision Variables:
workerHours[w,h]: Indicates if worker w is assigned hour h
workerTasks[w,t]: Indicates if worker w is assigned task t
startBlock[w,h]: Indicates if worker w is assigned a consecutive block of hours starting at hour h
endBlock[w,h]: Indicates if worker w is assigned a consecutive block of hours ending at hour h
unassignedTasks[t]: Indicates if task t is not assigned all workers

Constraints:
1) Available hours constraint. If a worker is assigned an hour, he must be available for that hour.
workerHours[w,h] <= availableHours[w,h] for each w for each h
2) Worker task skill constraint. If a worker is assigned a task, he must be skilled enough for that task.
workerTask[w,t] <= workerTaskSkills[w,t] for each w for each t
3) Task hour assignment constraint. If a worker is assigned a task, he must be assigned the task's hour.
workerTask[w,t] <= workerHours[w,taskHour[t]] for each w for each t
4) At most 1 task assigned for each worker for each hour
sum(workerTask[w,t] for each t if taskHour[t]=h) <= 1 for each w for each h
5) Start block constraint. If a worker w starts a block of consecutive worker-hour assignments at hour h, startBlock[w,h] = 1. If workerHours[w,0], then startBlock[w,0]
workerHours[w,h] + (1-workerHours[w,h-1]) - 1 <= startBlock[w,h] for each w for each h
workerHours[w,0] <= startBlock[w,0] for each w
6) End block constraint. If a worker w ends a block of consecutive worker-hour assignments at hour h, endBlock[w,h] = 1. If workerHours[w,H-1], then endBlock[w,H-1]
workerHours[w,h] + (1-workerHours[w,h+1]) - 1 <= endBlock[w,h] for each w for each h
workerHours[w,H-1] <= endBlock[w,H-1] for each w
7) Minimum block length constraint. If a worker starts a block at hour h, the worker must be assigned the at least {BMin} hours.
startBlock[w,h] <= workerHours[w,x] for each w for each h for each x in range(h,h+BMin)
8) Maximum block length constraint. If a worker starts a block at hour h, the worker must end within {BMax} hours.
startBlock[w,h] <= sum(endBlock[w,x] for x in range(h,h+BMax)) for each w for each h
9) Minimum Rest constraint. The minimum length of rest (block of consecutive worker-hour = 0) between the end of a block and the start of a block is {RMin}.
endBlock[w,h] <= 1-workerHours[w,x] for each w for each h for each x in range(h+1,h+RMin+1)
10) Unassigned Task constraint. If a task is not assigned to any worker, unassignedTasks[t] = 1.

Given the above context, provide the mathematical formulation for the new constraint mentioned below:
New Constraint:
11) Formulate a constraint such that worker A does not work from hours H1 to H2.

You are free to add parameters and decision variables.

New Parameters:
A: Worker affected from hours H1 to H2
H1: Start of hours where worker A does not work
H2: End of hours where worker A does not work

New Variables:
None

New Constraints:
workerHours[A,h] = 0 for each h in range(H1,H2)

*Current Code*
```python
from ortools.sat.python import cp_model
import pandas as pd
import sys
import os
sys.path.append(os.getcwd())
from readdata import read_input_json

# Read data
current_dir = os.getcwd()
# Parameters
input_data_file = os.path.join(current_dir, "gsp_input.json")
data = read_input_json(input_data_file)
W = data["W"]
H = data["H"]
T = data["T"]
availableHours = data["availableHours"]
taskHour = data["taskHour"]
workerTaskSkills = data["workerTaskSkills"]
BMin = data["BMin"]
BMax = data["BMax"]
RMin = data["RMin"]

# Start model object
model = cp_model.CpModel()

# Create Variables
workerHours = model.NewBoolVarSeries("workerHours", pd.Index([(w,h) for w in range(W) for h in range(H)]))
startBlock = model.NewBoolVarSeries("startBlock", pd.Index([(w,h) for w in range(W) for h in range(H)]))
endBlock = model.NewBoolVarSeries("endBlock", pd.Index([(w,h) for w in range(W) for h in range(H)]))
taskAssigned = model.NewBoolVarSeries("taskAssigned", pd.Index([(w,t) for w in range(W) for t in range(T)]))
unassignedTask = model.NewBoolVarSeries("unassignedTask", pd.Index([t for t in range(T)]))

# Main Constraints - start
## Constraint 1: Available hours constraint
for w in range(W):
    for h in range(H):
        model.Add(workerHours[w,h] <= availableHours[w,h])

## Constraint 2: Worker task skill constraint
for w in range(W):
    for t in range(T):
        model.Add(taskAssigned[w,t] <= workerTaskSkills[w,t])

## Constraint 3: Task hour assignment constraint
for w in range(W):
    for t in range(T):
        model.Add(taskAssigned[w,t] <= workerHours[w,taskHour[t]])

## Constraint 4: At most 1 task assigned for each worker for each hour
for w in range(W):
    for h in range(H):
        model.Add(sum(taskAssigned[w,t] for t in range(T) if taskHour[t] == h) <= 1)

## Constraint 5: startBlock[w,h] = 1 indicates the start of a consecutive assignment of workerHours for worker w starting at hour h.
for w in range(W):
    for h in range(1, H):
        model.Add(workerHours[w,h-1].Not() + workerHours[w,h] - 1 <= startBlock[w,h])
    
    model.Add(workerHours[w,0] <= startBlock[w,0])

## Constraint 6: endBlock[w,h] = 1 indicates the end of a consecutive assignment of workerHours for worker w ending at hour h.
for w in range(W):
    for h in range(0,H-1):
        model.Add(workerHours[w,h] + workerHours[w,h+1].Not() - 1 <= endBlock[w,h])
    
    model.Add(workerHours[w,H-1] <= endBlock[w,H-1])

## Constraint 7: Min Block Length constraint
for w in range(W):
    for h in range(H-BMin):
        for x in range(h, h+BMin):
            model.Add(startBlock[w,h] <= workerHours[w,x])

    for h in range(H-BMin, H):
        model.Add(startBlock[w,h] == 0)

## Constraint 8: Max Block Length Constraint
for w in range(W):
    for h in range(H-BMax-1):
        model.Add(startBlock[w,h] <= sum(workerHours[w,x].Not() for x in range(h+1,h+BMax+1)))

## Constraint 9: Minimum Rest constraint
for w in range(W):
    for h in range(H):
        for x in range(h+1, min(H, h+RMin+1)):
            model.Add(endBlock[w,h] <= workerHours[w,x].Not())

## Constraint 10: Unassigned task constraint
for t in range(T):
    model.Add(sum(taskAssigned[w,t] for w in range(W)) >= 1 - unassignedTask[t])
# Main Constraints - end

# Dynamic Constraints

# Objective
total_hours_assigned = sum(workerHours[w,h] for w in range(W) for h in range(H))
total_unassigned_tasks = sum(unassignedTask[t] for t in range(T))
UNASSIGNED_PENALTY = 1000
objective = total_hours_assigned + UNASSIGNED_PENALTY * total_unassigned_tasks
model.Minimize(objective)

# Run Solver
solver = cp_model.CpSolver()
solver.parameters.linearization_level = 0
solver.parameters.max_time_in_seconds = 180
solver.parameters.optimize_with_core = True
solver.parameters.num_search_workers = 1
solver.parameters.log_search_progress = True
status = solver.Solve(model)
```
*Coding Agent*
```python
from ortools.sat.python import cp_model
import pandas as pd
from readdata import read_input_json

# Global Parameters
input_data_file = "simple_input.json"
data = read_input_json(input_data_file)
W = data["W"]
H = data["H"]
T = data["T"]
availableHours = data["availableHours"]
taskHour = data["taskHour"]
workerTaskSkills = data["workerTaskSkills"]
BMin = data["BMin"]
BMax = data["BMax"]
RMin = data["RMin"]

A = data["A"]
H1 = data["H1"]
H2 = data["H2"]
# Global Parameters - end

# Start model object
model = cp_model.CpModel()

# Global Variables
workerHours = model.NewBoolVarSeries("workerHours", pd.Index([(w,h) for w in range(W) for h in range(H)]))
startBlock = model.NewBoolVarSeries("startBlock", pd.Index([(w,h) for w in range(W) for h in range(H)]))
endBlock = model.NewBoolVarSeries("endBlock", pd.Index([(w,h) for w in range(W) for h in range(H)]))
taskAssigned = model.NewBoolVarSeries("taskAssigned", pd.Index([(w,t) for w in range(W) for t in range(T)]))
unassignedTask = model.NewBoolVarSeries("unassignedTask", pd.Index([t for t in range(T)]))

None
# Global Variables - end

# Main Constraints - start
## Constraint: 1 
"""
Desc: Skill prequisite constraint: If a task is assigned to a worker, the worker-task skills must be compatible. If taskAssigned[w,t] = 1, then workerTaskSkills[w,t] = 1
Type: Main
Parameters: None
Variables: None
"""
### Code:
for w in range(W):
    for t in range(T):
        model.Add(taskAssigned[w,t] <= workerTaskSkills[w,t])

## Constraint: 2
"""
Desc: Time prerequisite constraint: If a task is assigned to a worker, the worker must be assigned to the task's hour. If taskAssigned[w,t] = 1, then workerHours[w,taskHour[t]] = 1
Type: Main
Parameters: None
Variables: None
"""
### Code:
for w in range(W):
    for t in range(T):
        model.Add(taskAssigned[w,t] <= workerHours[w,taskHour[t]])

## Constraint: 3 
"""
Desc: No multi-tasking constraint: A worker can only be assigned at most 1 task per hour. A worker can work at most 1 task per hour
Type: Main
Parameters: None
Variables: None
"""
### Code:
for w in range(W):
    for h in range(H):
        model.Add(sum(taskAssigned[w,t] for t in range(T) if taskHour[t] == h) <= 1)

## Constraint: 4
"""
Desc: Time availability constriant: If a worker is assigned an hour, the worker must be available for that hour. If workerHours[w,h] = 1, then availableHours[w,h] = 1
Type: Main
Parameters: None
Variables: None
"""
### Code:
for w in range(W):
    for h in range(H):
        model.Add(workerHours[w,h] <= availableHours[w,h])

## Constraint: 5
"""
Desc: startBlock[w,h] = 1 indicates the start of a consecutive assignment of workerHours for worker w starting at hour h. If workerHours[w,h-1] = 0 and workerHours[w,h] = 1, then startBlock[w,h] = 1. If workerHours[w,0] = 1, then startBlock[w,0] = 1.
Type: Main
Parameters: None
Variables: None
"""
### Code:
for w in range(W):
    for h in range(1, H):
        model.Add(workerHours[w,h-1].Not() + workerHours[w,h] - 1 <= startBlock[w,h])
    
    model.Add(workerHours[w,0] <= startBlock[w,0])

## Constraint: 6 
"""
Desc: endBlock[w,h] = 1 indicates the end of a consecutive assignment of workerHours for worker w ending at hour h. If workerHours[w,h-1] = 1 and workerHours[w,h] = 0, then endBlock[w,h-1] = 1. If workerHours[w,H-1] = 1, then endBlock[w,H-1] = 1.
Type: Main
Parameters: None
Variables: None
"""
### Code:
for w in range(W):
    for h in range(1, H):
        model.Add(workerHours[w,h-1] + workerHours[w,h].Not() - 1 <= endBlock[w,h-1])
    
    model.Add(workerHours[w,H-1] <= endBlock[w,H-1])

## Constraint: 7 
"""
Desc: If a task is not assigned to a single worker (if taskAssigned[w,t] = 0 for all w), then unassignedTask[t] = 1. If unassignedTask[t] = 0, then sum(taskAssigned[w,t] for all w) >= 1
Type: Main
Parameters: None 
Variables: None
"""
### Code:
for t in range(T):
    model.Add(sum(taskAssigned[w,t] for w in range(W)) >= 1 - unassignedTask[t])

## Constraint: 8 
"""
Desc: Min Block Length constraint: Minimum length of consecutive hours assigned to a worker is {BMin}. If startBlock[w,h] = 1, then workerHours[w,h] = 1 and workerHours[w,h+1] = 1 and ... workerHours[w,h+BMin-1] = 1
Type: Main
Parameters: None
Variables: None
"""
### Code:
for w in range(W):
    for h in range(H-BMin):
        for x in range(h, h+BMin):
            model.Add(startBlock[w,h] <= workerHours[w,x])

    for h in range(H-BMin, H):
        model.Add(startBlock[w,h] == 0)

## Constraint: 9 
"""
Desc: Max Block Length Constraint: Maximum length of consecutive hours assigned to a worker is {BMax}. If startBlock[w,h] = 1, then NOT (workerHours[w,h+1]=1 and ... workerHours[w,h+BMax]=1). If startBlock[w,h] = 1, then workerHours[w,h+1]=0 or ... workerHours[w,h+BMax]=0
Type: Main
Parameters: None
Variables: None
"""
### Code:
for w in range(W):
    for h in range(H-BMax-1):
        model.Add(startBlock[w,h] <= sum(workerHours[w,x].Not() for x in range(h+1,h+BMax+1)))

## Constraint: 10 
"""
Desc: Minimum Rest constraint: Minimum length of consecutive non-working hours after hour h for worker w = {RMin} if endBlock[w,h] = 1. If endBlock[w,h], then (workerHours[w,h+1]=0 and workerHours[w,h+2]=0 and ... workerHours[w,h+RMin]=0)
Type: Main
Parameters: None
Variables: None
"""
### Code:
for w in range(W):
    for h in range(H):
        for x in range(h+1, min(H, h+RMin+1)):
            model.Add(endBlock[w,h] <= workerHours[w,x].Not())

## Constraint: 11

### Code:

for h in range(H1, H2+1):
    model.Add(workerHours[A,h] == 0)

# Objective
total_hours_assigned = sum(workerHours[w,h] for w in range(W) for h in range(H))
total_unassigned_tasks = sum(unassignedTask[t] for t in range(T))
UNASSIGNED_PENALTY = 1000
objective = total_hours_assigned + UNASSIGNED_PENALTY * total_unassigned_tasks
model.Minimize(objective)

# Run Solver
solver = cp_model.CpSolver()
solver.parameters.linearization_level = 0
solver.parameters.max_time_in_seconds = 60 * 3 
solver.parameters.optimize_with_core = True
solver.parameters.num_search_workers = 1
solver.parameters.log_search_progress = True
status = solver.Solve(model)
```

Your turn:
Problem:
You are tasked to solve a nurse rostering problem. 
In this problem, there are {N} nurses, {D} days, {S} shifts and {T} slots.
Each worker can be a shift for each day.
Each shift takes a certain number of hours, specified in {shiftHours}.
Each shift fills either a Morning(AM)(Slot 0), Afternoon(PM)(Slot 1), or Night(ND)(Slot 2) slot.
There is a demand requirement for each day for each slot, where we want to assign the nurses their shifts as close as possible to fulfil this demand.
The main task is to assign nurses a shift for each day (or no shift if the nurse is not working for that day), while satisfying some constraints and minimizing an objective.
The objective is to minimize the following:
surplus + shortfall
surplus is the total number of nurses assigned to a slot more than the specified demand for that day and slot.
shortfall is the total number of nurses assigned to a slot less than the specified demand for that day and slot.

Parameters:
N: Number of nurses
D: Number of days
S: Number of shifts
T: Number of slots
availableShifts[n,d,s]: Indicates if nurse n is available to be assigned day d and slot s (1 if available, otherwise 0)
shiftSlot[s] returns the slot number for shift s. (Eg. 7am to 11am is a Morning shift(AM), so it has a value of 0 (Slot 0))
shiftHours[s] = Integer value for the number of hours in shift s. (Eg. 7am to 11am shift has a value of 4)
demandSlot[d,t] = Integer value of the demand required specified per day d per slot t
restDays[n,d]: Indicates if nurse n is resting (Not available) on day d. 
MinHours = data["MinHours"]
MaxHours = data["MaxHours"]
MaxWorkingDays = data["MaxWorkingDays"]

Decision Variables:
nurseDayShift[n,d,s]: Indicates if nurse n is assigned day d and shift s
nurseDaySlot[n,d,t]: Indicates if nurse n is assigned a shift in day d slot t
surplus[d,t] = Integer value of the surplus nurses assigned on day d slot t. (If X more nurses are assigned to day d slot t than specified in demandSlot[d,t], return X, otherwise 0)
shortfall = Integer value of the shortfall nurses assigned on day d slot t. (If X less nurses are assigned to day d slot t than specified in demandSlot[d,t], return X, otherwise 0)

Constraints:
1) Available shift constraint. If nurse n is assigned day d and shift s, she must be available on day d shift s.
nurseDayShift[n,d,s] <= availableShifts[n,d,s] for each n for each d for each s
2) Max 1 shift per day constraint. At most 1 shift per day constraint
sum(nurseDayShift[n,d,s] for each s) <= 1, for each n for each s
3) Shift to slot assignment constriant. If a nurse n is assigned day d shift s, then nurseDaySlot[n,d,shiftSlot[s]] = 1
nurseDaySlot[n,d,t] <= sum(nurseDayShift[n,d,s] for each s if shiftSlot[s] == t)
4) Demand Requiremented Constraint. Number of nurses assigned for each day d for each slot t == demandSlot[d,t] + surplus[d,t] - shortfall[d,t]
sum(nurseDaySlot[n,d,t] for each n) == demandSlot[d,t] + surplus[d,t] - shortfall[d,t]
5) Minimum Hours Constraint. Each nurse must work at at least {MinHours}
sum(shiftHours[s] * nurseDayShift[n,d,s] for each d for each s) >= MinHours, for each n
6) Maximum Hours Constraint: Each nurse must work at most {MaxHours}
sum(shiftHours[s] * nurseDayShift[n,d,s] for each d for each s) <= MaxHours, for each n
7) Rest Days Constraint. If a nurse n is assigned any shift on day d, she must not be on a rest.
nurseDayShift[n,d,s] <= 1 - restDays[n,d] for each n for each d for each s
8) Maximum number of working days per week constraint. Each nurse must work at most (MaxWorkingDays) per week.
sum(nurseDayShift[n,w*7 + d,s] for d in range(7) for each s) <= MaxWorkingDays, for each n for each week w
9) No consecutive Night(ND) shift - Morning (AM) shift assignment for any nurse constraint.
nurseDaySlot[n,d,ND] + nurseDaySlot[n,d+1,AM] <= 1, for each n for each d

New Constraint
Nurse K is not available to work on day D1.

### Parameters ###
K: Nurse affected on day D1.
D1: Nurse K cannot work on this day.

### Variables ###
None

### Constraints ###
nurseDayShift[K,D1,s] = 0 for each s

Current code:
```python
from ortools.sat.python import cp_model
import pandas as pd
import numpy as np
import sys
import os
sys.path.append(os.getcwd())
from readdata import read_input_json

# Read data
current_dir = os.getcwd()
input_file = os.path.join(current_dir, "nsp_input.json")
data = read_input_json(input_file)

# Parameters
N = data["N"] # Num nurses
D = data["D"] # Num days
S = data["S"] # Num Shifts. A shift can be 7am to 11am, 9am to 1pm, 8pm to 7am, etc.
T = data["T"] # Num Slots: AM = 0, PM = 1, ND = 2 (Morning, Afternoon Night). A shift is classified under one of these 3 slots.
AM,PM,ND = 0,1,2
availableShifts = data["availableShifts"] # availableShifts[n,d,s] returns a boolean if nurse n is available for day d and shift s.
shiftSlot = data["shiftSlot"] # shiftSlot[s] returns the slot number for shift s. (7am to 11am is a Morning shift(AM), so it has a value of 0)
shiftHours = data["shiftHours"] # shiftHours[s] returns the number of hours in shift s. (7am to 11am shift has a value of 4)
demandSlot = data["demandSlot"] # Demand is specified per day per slot in demandSlot[d,t]
restDays = data["restDays"] # restDays[n,d]=1 implies that availableShifts[n,d,s] = 0 for all s
MinHours = data["MinHours"]
MaxHours = data["MaxHours"]
MaxWorkingDays = data["MaxWorkingDays"]

# Start model object
model = cp_model.CpModel()

# Variables
nurseDayShift = model.NewBoolVarSeries("nurseDayShift", pd.Index([(n,d,s) for n in range(N) for d in range(D) for s in range(S)]))
nurseDaySlot = model.NewBoolVarSeries("nurseDaySlot", pd.Index([(n,d,t) for n in range(N) for d in range(D) for t in range(T)]))
surplus = model.NewIntVarSeries("surplus", pd.Index([(d,t) for d in range(D) for t in range(T)]), 0, 3)
shortfall = model.NewIntVarSeries("shortfall", pd.Index([(d,t) for d in range(D) for t in range(T)]), 0, 3)

# Constraints
## Available shift constraint
for n in range(N):
    for d in range(D):
        for s in range(S):
            model.Add(nurseDayShift[n,d,s] <= availableShifts[n,d,s])

## At most 1 shift per day constraint
for n in range(N):
    for d in range(D):
        model.Add(sum(nurseDayShift[n,d,s] for s in range(S)) <= 1)

## Nurse Shift Slot assignment Constraint
for n in range(N):
    for d in range(D):
        for t in range(T):
            model.Add(nurseDaySlot[n,d,t] <= sum(nurseDayShift[n,d,s] for s in range(S) if shiftSlot[s] == t))

## Demand Constraint: Number of nurses assigned for each day d for each slot t == demandSlot[d,t] + surplus[d,t] - shortfall[d,t]
for d in range(D):
    for t in range(T):
        model.Add(sum(nurseDaySlot[n,d,t] for n in range(N)) == demandSlot[d,t] + surplus[d,t] - shortfall[d,t])

## Min Hours Constraint
for n in range(N):
    model.Add(sum(nurseDayShift[n,d,s] * shiftHours[s] for d in range(D) for s in range(S)) >= MinHours)

## Max Hours Constraint
for n in range(N):
    model.Add(sum(nurseDayShift[n,d,s] * shiftHours[s] for d in range(D) for s in range(S)) <= MaxHours)

## Rest Day Constriant
for n in range(N):
    for d in range(D):
        for s in range(S):
            model.Add(nurseDayShift[n,d,s] <= 1- restDays[n,d])

## MaxWorkingDays = 5 per week
NumWeeks = D // 7
for n in range(N):
    for w in range(NumWeeks):
        model.Add(sum(nurseDayShift[n,w*7 + d,s] for d in range(7) for s in range(S)) <= MaxWorkingDays)

## No Night-Morning Shift pattern Constraint
## If a nurseDaySlot[n,d,ND] = 1, then nurseDaySlot[n,d+1,AM] = 0
for n in range(N):
    for d in range(D-1):
        model.Add(nurseDaySlot[n,d,ND] <= 1 - nurseDaySlot[n,d+1,AM])

## New Constraints

# Objective
demand_mismatch = sum(surplus[d,t] + shortfall[d,t] for d in range(D) for t in range(T))
model.Minimize(demand_mismatch)

# Run Solver
solver = cp_model.CpSolver()
solver.parameters.linearization_level = 0
solver.parameters.max_time_in_seconds = 60 * 2
solver.parameters.optimize_with_core = True
solver.parameters.num_search_workers = 16
solver.parameters.log_search_progress = True
status = solver.Solve(model)
 ```

\end{verbatim}
\end{small}